%% file: ClimbingCap_CVPR2025.tex
\documentclass[10pt,twocolumn,letterpaper]{article}

\usepackage{cvpr}              % To produce the CAMERA-READY version
\usepackage{graphicx}
\usepackage{amsmath}
\usepackage{amssymb}
\usepackage{booktabs}
\usepackage{stfloats}
\usepackage{color}
\usepackage{times}
\input{preamble}
\newcommand{\PAR}[1]{\vskip3pt \noindent{\bf #1~}}
\definecolor{cvprblue}{rgb}{0.21,0.49,0.74}
\usepackage[pagebackref,breaklinks,colorlinks,linkcolor=cvprblue,citecolor=cvprblue,urlcolor=cvprblue]{hyperref}
\usepackage{overpic}
\usepackage{bm}
\usepackage{tabu}
\usepackage{tabularx}
\usepackage{bbding}
\usepackage{multicol}
\usepackage{multirow}
\usepackage{scalerel,stackengine}
\usepackage[accsupp]{axessibility}
\usepackage{array}
\usepackage[capitalize]{cleveref}
\crefname{section}{Sec.}{Secs.}
\Crefname{section}{Section}{Sections}
\Crefname{table}{Table}{Tables}
\crefname{table}{Tab.}{Tabs.}
\usepackage{algorithm}
\usepackage{algorithmic}

\usepackage[capitalize]{cleveref}
\crefname{section}{Sec.}{Secs.}
\Crefname{section}{Section}{Sections}
\Crefname{table}{Table}{Tables}
\crefname{table}{Tab.}{Tabs.}

\def\paperID{6231} % *** Enter the CVPR Paper ID here
\def\confName{CVPR}
\def\confYear{2025}

\title{
ClimbingCap: Multi-Modal Dataset and Method for Rock Climbing \\ in World Coordinate
\vspace{-4mm}
}

\author{Ming Yan$^{1,2,3\ast}$\hspace{4mm} Xincheng Lin$^{1,3\ast}$\hspace{4mm} Yuhua Luo$^{1,3}$\hspace{4mm} Shuqi Fan$^{1,3}$\hspace{4mm} Yudi Dai$^{7}$\hspace{4mm} Qixin Zhong$^{4}$\hspace{4mm} \\ Lincai Zhong$^{5}$\hspace{4mm} Yuexin Ma$^6$\hspace{4mm} Lan Xu$^6$\hspace{4mm} Chenglu Wen$^{1,3}$\hspace{4mm} Siqi Shen$^{1,3\dagger}$\hspace{4mm} Cheng Wang$^{1,3}$ \vspace{+1mm}
\\
$^1$Fujian Key Laboratory of Sensing and Computing for Smart Cities, Xiamen University\\
$^2$National Institute for Data Science in Health and Medicine, Xiamen University\\
$^3$Key Laboratory of Multimedia Trusted Perception and Efficient Computing,\\ Ministry of Education of China, School of Informatics, Xiamen University\\
$^4$China National Climbing Team \hspace{2mm}$^5$Ningbo Sports Work Training Team\\
$^6$ShanghaiTech University\hspace{2mm}$^7$ETH AI Center, ETH Zürich\\
{}
\vspace{-18mm}
\and
\\
{}
}

\begin{document}
\twocolumn[{%
\renewcommand\twocolumn[1][!htb]{#1}%
\maketitle
\vspace{-10mm}
\begin{center}
    \centering
    \includegraphics[width=1\linewidth]{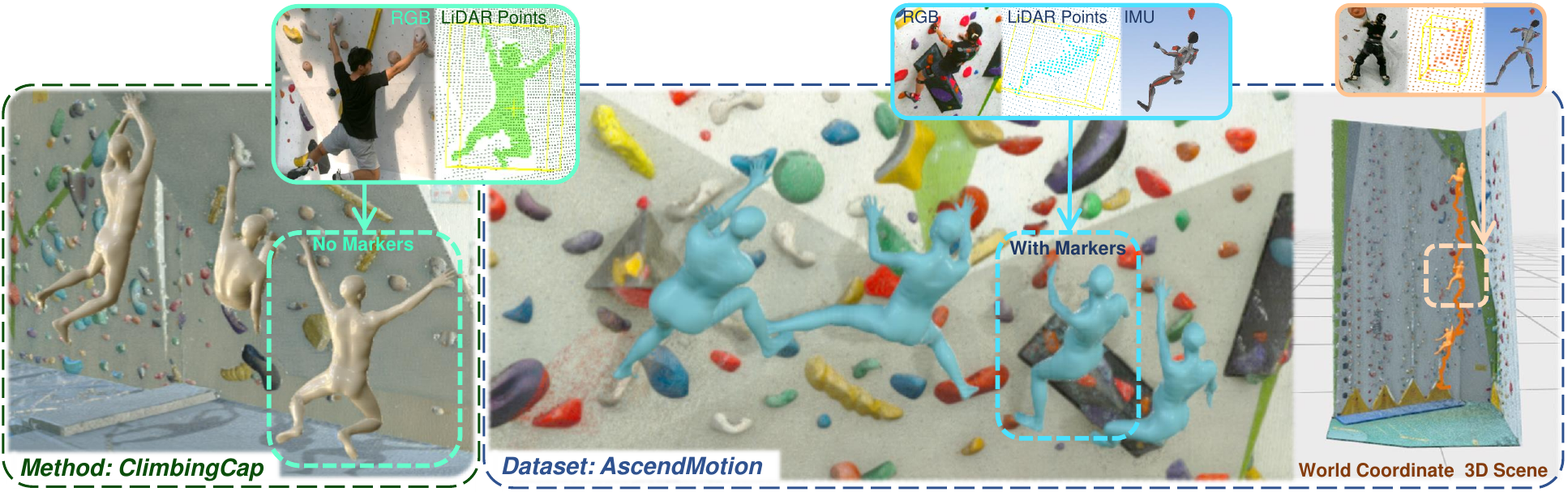}
    \vspace{-7mm}
    \captionof{figure}{\textbf{Overview.} To address the challenging problem of global climbing motion recovery, we collect the dataset \textbf{AscendMotion}, using LiDAR, RGB camera and Inertial Measurement Unit (IMU) motion capture system with accurate motion labels and global trajectories (the \textcolor[rgb]{0.368, 0.6, 0.615}{blue} and \textcolor[rgb]{0.749, 0.506, 0.275}{orange} human bodies in right side of the figure represent labeled motions, and the \textcolor[rgb]{0.882, 0.674, 0.435}{orange curve} represents the motion trajectory in the world coordinate.). Meanwhile, we propose \textbf{ClimbingCap}, a global climbing motion capturing method in world coordinate. As shown in the \textcolor[rgb]{0.4, 1, 0.796}{left} part of this figure, it uses both image and LiDAR point cloud to recover human motions. }
%    \captionof{figure}{\textbf{Overview.} In order to address the challenging problem of pose and global motion recovery in the world coordinate system, we collect the dataset \textbf{AscendMotion}, using LiDAR, RGB camera and IMU (Inertial Measuring Unit) motion capture system for precise annotation in the world coordinate system to obtain accurate motion labels and global trajectories (the \textcolor[rgb]{0.368, 0.6, 0.615}{blue} and \textcolor[rgb]{0.749, 0.506, 0.275}{orange} human bodies in the right figure represent the global recovery results, and the \textcolor[rgb]{0.882, 0.674, 0.435}{orange curve} represents the world coordinate motion trajectory.). Meanwhile, we propose \textbf{ClimbingCap}, a global HMR framework that uses both image and point cloud information to improve human motion recovery performance. As shown in the \textcolor[rgb]{0.4, 1, 0.796}{left} figure, ClimbingCap can recover global human motion from in-the-wild data.}

    \label{fig:gallery_climbing}
    % \vspace{-1mm}
\end{center}%
}]

% \begin{figure*}[!htb]
%     \centering
%     \includegraphics[width=\textwidth]{Pics/Climbing_FrameWork.pdf}
%     \caption{CIME4D looks handsome}
%     \label{fig: facade picture2}
% \end{figure*}

\footnotetext{$^\ast$ Equal contribution.}
\footnotetext{$^\dagger$ Corresponding author.}

%%%%%%%%% ABSTRACT
\input{Secs/abstract.tex}

%%%%%%%%% BODY TEXT
\vspace{-5mm}
\input{Secs/Introduction}

\label{sec:intro}

\input{Secs/relatedWork}

\label{sec:relatedWork}

\input{Secs/Method_ClimbingCap}

\label{sec:Method_ClimbingCap}

\input{Secs/Dataset_AscendMotion}
\label{sec:Dataset_AscendMotion}

\input{Secs/Experiment}

\input{Secs/Conclusion}
{\small
\bibliographystyle{ieee_fullname}
\bibliography{egbib}
}

\end{document}

%% file: preamble.tex
\usepackage[dvipsnames]{xcolor}

%% file: Secs/Abstract.tex
\begin{abstract}
%\vspace{-4mm}

Human Motion Recovery (HMR) research mainly focuses on ground-based motions such as running. The study on capturing climbing motion, an off-ground motion, is sparse. This is partly due to the limited availability of climbing motion datasets, especially large-scale and challenging 3D labeled datasets. To address the insufficiency of climbing motion datasets, we collect \textbf{AscendMotion}, a large-scale well-annotated, and challenging climbing motion dataset. It consists of 412k RGB, LiDAR frames, and IMU measurements, including the challenging climbing motions of 22 skilled climbing coaches across 12 different rock walls. Capturing the climbing motions is challenging as it requires precise recovery of not only the complex pose but also the global position of climbers. Although multiple global HMR methods have been proposed, they cannot faithfully capture climbing motions. To address the limitations of HMR methods for climbing, we propose \textbf{ClimbingCap}, a motion recovery method that reconstructs continuous 3D human climbing motion in a global coordinate system. One key insight is to use the RGB and LiDAR modalities to separately reconstruct motions in camera coordinates and global coordinates and to optimize them jointly. We demonstrate the quality of the AscendMotion dataset and present promising results from ClimbingCap. The AscendMotion dataset and source code release publicly at \href{http://www.lidarhumanmotion.net/climbingcap/}{http://www.lidarhumanmotion.net/climbingcap/}

\end{abstract}

%% file: Secs/Introduction.tex
\section{Introduction}
\label{sec:intro}

It is challenging for global Human Motion Recovery (HMR)~\cite{Martinez17,alldieck2017optical,SMPL2015,Zhou16a,OpenPose} due to the complexity of human poses and dynamic interaction between humans and their environments. Moreover, global HMR must maintain consistency in world coordinates to ensure authenticity and physical feasibility. The research community has proposed various methods to estimate human poses from different sensors such as RGB images~\cite{SPIN_ICCV2019,MonoPerfCap,DeepCap_CVPR2020,challencap,LiveCap2019tog}, LiDAR point clouds~\cite{lidarcap,zhang2024lidarcapv2, ren2024livehps}, and inertial measurement units (IMUs)~\cite{DIP,PIP}. Most of the studies~\cite{Bregler1998TrackingPW,Vlasic2007PracticalMC,woltring1974new,kaufmann2023emdb,dai2024hisc4d,RICH,zhang2024lidarcapv2,ren2024livehps,lin2024hmpear,dai2023sloper4d,yan2024reli11d} focus on recovering ground-based motions such as running, interacting, and dancing.

Unlike ground-based motions such as running, climbing is an activity performed off-ground, where climbers use hands and feet to ascend holds or walls. As an important sport and recreation, the climbing race has become an official event in the Olympic Games. Yet, research on climbing motion capture~\cite{ViconForceClimbing,exemPose,MRClimbing17,ClimbingMR,footVR} is sparse, partly due to the limited availability of climbing datasets. There are only two publicly available climbing datasets: SPEED21~\cite{elias2021speed21} and CIMI4D~\cite{yan2023cimi4d}. SPEED21 is a 2D dataset. Although CIMI4D is 3D, it consists of many trivial climbing motions from casual climbers. Moreover, the size of the two datasets is small. As a result, the research community lacks an in-depth understanding of challenging climbing motions.

%Unlike ground-based motions such as running, climbing is an activity performed off-ground, where climbers use hands and feet to ascend holds or walls. It is an important sport and recreation~\cite{ViconForceClimbing,exemPose,MRClimbing17,ClimbingMR,footVR}. The climbing race is an official event in the Olympic Games. Despite its popularity, research on climbing motion capturing, is sparse. This situation is partly due to the limited availability of climbing datasets. There are only two publicly available Climbing datasets: SPEED21~\cite{elias2021speed21} and CIMI4D~\cite{yan2023cimi4d}. SPEED21 is a 2D dataset, whereas CIMI4D is a dataset that includes 3D data. The size of these two datasets is small, and the 3D climbing motions consist of many amateur and trivial climbing motions. As a result, the research community lacks an in-depth understanding of the challenging climbing motions. 

We collect \textbf{AscendMotion}, a large-scale and challenging climbing motion multi-modal dataset, to address the data insufficiency. AscendMotion dataset includes the motions of 22 skilled climbers climbing 12 different rock walls. It includes rich modalities from hardware time-synchronized RGB, LiDAR point clouds, and Inertial Measurement Unit(IMU) MoCap system. Moreover, it consists of the global trajectories of climbers, which are important for understanding human motions in world coordinates. We also combine automatic annotation with manual refinement to ensure the accuracy of the motion annotation. 
% We demonstrate the quality of our dataset through experiments.

Not only does climbing motion challenge humans, but capturing climbing motion is also challenging for researchers. Capturing climbing motion requires precise estimation of the complex pose resulting from human-scene interactions, as well as the global localization of climbers within scenes. Existing research dedicated to climbing~\cite{Bregler1998TrackingPW,Vlasic2007PracticalMC,woltring1974new} uses standard 2D HMR (e.g. OpenPose~\cite{OpenPose}) for climbing motions, which is suboptimal. Most existing HMR methods~\cite{shen2024gvhmr,goel2023humans,shin2024wham,li2025coin,ye2023slahmr,sun2023trace} focus on recovering ground-based human motions through RGB imagery. The estimation results of these methods cannot be rigidly transformed among camera coordinates and global coordinate systems, due to the inherent ambiguity between coordinate systems. Moreover, global HMR methods often accumulate errors during long-term sequence recovery, affecting the accuracy of global trajectories and postures, especially in climbing movements, as we show through experiments.

%Not only does climbing motion challenge humans, but capturing climbing motion is also challenging for researchers. Capturing climbing motion requires precise estimation of the complex pose resulting from human-scene interactions in the camera coordinate, as well as the global coordinates of climbers and scenes.
%Existing research dedicat to climbing~\cite{Bregler1998TrackingPW,Vlasic2007PracticalMC,woltring1974new} use standard 2D pose estimation algorithms (e.g., OpenPose\cite{OpenPose}) for climbing motions, which are sub-optimal. Most existing HMR methods focus on recovering human motion~\cite{shen2024gvhmr,goel2023humans,shin2024wham,li2025coin,ye2023decoupling,sun2023trace}. The estimation results of these methods cannot be rigidly transformed into other coordinate systems due to the inherent ambiguity between coordinate systems. Moreover, global HMR methods often accumulate errors during long-term sequence recovery, affecting the accuracy of global trajectories and postures, especially in climbing movements as we show in the experiments.

%Because of these challenges and the lack of datasets, there are no HMR algorithms that work well for climbing motions. 
To tackle the above challenges and fill the gap in the motion capture community for climbing motions, we introduce \textbf{ClimbingCap}, an HMR method that reconstructs continuous 3D human climbing motion in both camera and global coordinate systems (see Fig.\ref{fig:gallery_climbing}). We adopt a trilogy in ClimbingCap: separate coordinate decoding, post-processing, and semi-supervised training. For the separate coordinate decoding stage, ClimbingCap uses image and point cloud modality to estimate poses in the camera and the global coordinate systems, respectively. The post-processing stage accurately ensures the consistency of motions between the two coordinate systems. In the semi-supervised training stage, a teacher-student training method is adopted to make use of easily obtainable unlabeled climbing data. Through these innovations, we can handle complex climbing movements effectively.
To summarize, our main contributions include:
\begin{itemize}
    \item We collect the dataset AscendMotion, which is more comprehensive than existing datasets. We demonstrate the quality of the dataset and its poses are challenging. 
    \item  We propose ClimbingCap, which is a multimodality global HMR method for rock climbing. 
    \item We conduct extensive experiments on evaluate the ClimbingCap in multiple datasets with various state-of-the-art methods. The experimental results show ClimbingCap perform better than all the methods for climbing motions. 
    
\end{itemize}

%% file: Secs/relatedWork.tex
\section{Related Work}\label{sec:related}

\PAR{Camera-Space HMR.} Camera-space HMR methods~\cite{hmrKanazawa17, VIBE_CVPR2020, li2022cliff, BEV, ROMP, PARE_ICCV2021, chen2021sportscap, Bogo2016KeepIS, SIMPX, MAED, pifu, pifuhd, ICON, wang2023nemo, GFPose2023, MotionBert2023, TORE2023, HUMOR_ICCV2021, Su2020RobustFusionHV,yang2021s3,Patel2020TailorNetPC} predict human motion in independent camera coordinate system for each frame of the input video. Although camera-space methods can predict accurate human poses, they cannot effectively reflect the orientation and trajectory of the human body in world coordinates.

%\PAR{Camera-Space Human Motion Recovery.}Most of the recent human motion recovery methods are based on parameterized human models for training and inference, such as SMPL ~\cite{SMPL2015}. The model parameters obtained from the network are then converted into a human body mesh, which can be further rendered back into a 2D image via the camera intrinsic. Camera-Space Human Motion Recovery Method~\cite{hmrKanazawa17, VIBE_CVPR2020, li2022cliff, BEV, ROMP, PARE_ICCV2021} refers to predicting human motion in independent camera coordinate system for each frame of the input video. Although Camera-Space methods can predict accurate human poses, their predictions are located in the camera coordinate system. It cannot effectively reflect the orientation and trajectory of the human body in the world space.

\input{tables/Dataset_Compare}

% Recent studies have achieved accurate results in World-Space, and most of the recent methods are based on two stages: 1.Tracking: based on target detection to obtain the 2d-joints of the human body, and based on the pre-trained human pose estimation model to obtain the human body pose in the camera space. 2.Global Orientation and Trajectory Optimization: based on the results of first stage, further optimization based on the global information provided by the camera and the scene.

\PAR{World-Grounded HMR.} 
Recent studies~\cite{kocabas2024pace,yuan2022glamr,ye2023slahmr,shen2024gvhmr,shin2024wham,sun2023trace,li2025coin} have achieved remarkable progress in world coordinates. Most of them are two-stage methods. The first stage estimates human motion in camera coordinates, and the second stage optimizes the motion in world coordinates with the scene information. Most global HMR methods focus on ground-based motions (e.g., walking) that are located on the ground. They cannot accurately model climbing motions, which is an off-ground motion. Our method, ClimbingCap, is a global HMR method developed for recovering rock climbing motions.

\PAR{HMR Dataset.} We have compared the AscendMotion dataset with multiple recent datasets in Table~\cref{tab:data_compare}. Most of the datasets focus on ground-based human daily activities or sports. There exist only two public available climbing motion datasets. SPEED21~\cite{elias2021speed21} collects 2D motions of climbers from sports videos. CIMI4D~\cite{yan2023cimi4d} contains 3D motions and global trajectories of climbers. Our climbing dataset, AscendMotion (344 minutes labeled data and 441 minutes unlabeled data, 412k frames), is significantly larger than both SPEED21 (21 minutes, 46k frames) and CIMI4D (120 minutes, 180k frames). Moreover, the motions in our dataset are more challenging than those in CIMI4D, as they come from climbing coaches who are more skilled than the casual climbers in CIMI4D.

\PAR{Climbing Motion Recovery.}
In recent years, the increasing popularity of climbing, coupled with the inclusion of rock climbing races in the Olympic Games, has led to growing interest in the capture and analysis of climbing motions~\cite{ClimbingSensorSurvey2022,replicating,cha2015analysis,richter2020human,beltran2023climbing}. Research on climbing MoCap methods remains limited. ~\cite{pandurevic2022analysis} employs OpenPose~\cite{OpenPose} for climbing pose estimation. ~\cite{reveret20203d} performs RGB-based method for local pose and estimates climbing position through a marker-based method.

%% file: tables/Dataset_Compare.tex
\begin{table*}[!t]
     % \vspace{-4mm}
     \centering
     \resizebox{1\linewidth}{!}{
     \begin{tabular}{lcccccccccccc}
        \toprule[1pt]
        
        \multirow{2}{*}{Dataset} & \multicolumn{3}{c}{Sensor Modalities} & \multirow{2}{*}{Global} & \multirow{2}{*}{Frames} & \multirow{2}{*}{3D Scene} & \multirow{2}{*}{Motion} & \multirow{2}{*}{Real/} & \multirow{2}{*}{Seqs} & \multirow{2}{*}{Subjects} & \multirow{2}{*}{Career}  \\

        \cline{2-4}
        
          & RGB & MoCap & LiDAR    & Trajectory &  &  & &Synthetic &  & &    \\
        
        \midrule
        
         % LiDARHuman26M\cite{lidarcap} &   \CheckmarkBold    &    IMU    &   \CheckmarkBold  & - & 184k &  - &   Daliy &   Real  & 20 & 13  \\
        % HSC4D~\cite{HSC4D} &   -    &  IMU     &    \CheckmarkBold          &    \CheckmarkBold    &   10k   &  \CheckmarkBold & Daliy & Real & 8 & 1   \\
        SLOPER4D~\cite{dai2023sloper4d} &    \CheckmarkBold   &     IMU  &    \CheckmarkBold          &     \CheckmarkBold   &   100k   &  \CheckmarkBold & Daliy & Real  & 15 & 12 &  Normal \\
   
        % LIPD~\cite{ren2023lidar} &   \CheckmarkBold    &   IMU    &    \CheckmarkBold          &    \CheckmarkBold    &  -    &  - & Obvious & Real  & 10 & 6   \\

        % mRI~\cite{an2022mri} &    \CheckmarkBold   &    IMU   &    mmWave    &   -   &      -  &   160k   &  - & Daliy & Real    \\
        % mmBody~\cite{chen2022mmbody} &    \CheckmarkBold   &    Optical   &    mmWave       &   -   &     -   &   200k   &  - & Daliy & Real     \\

        % mmMesh~\cite{xue2021mmmesh} &    \CheckmarkBold   &  Optical     &    mmWave    &   -   &    -     &   60k   & -  & Daliy & Real     \\

        % LiCamPose\cite{cong2022weakly} &   \CheckmarkBold    &   IMU    &    \CheckmarkBold          &    -    &  9k    &  - & Daliy & Real   & - & -   \\

        EMDB\cite{kaufmann2023emdb} &   \CheckmarkBold    &   EM    &    -          &    \CheckmarkBold    &  105k    &  - & Daliy & Real  & 81 & 10 & Normal   \\

        % SMART\cite{chen2021sportscap} &   \CheckmarkBold    &   -    &    -          &    -    &  110k    &  - & Sports & Real  & 640 & -   \\
        
        % X-Avatar\cite{shen2023x} &   \CheckmarkBold    &   -    &    -          &    -    &  35k    &  \CheckmarkBold & Daliy & Real  & 233 & 20   \\

        BEHAVE~\cite{bhatnagar2022behave} &  \CheckmarkBold     &    -    &    -          &    \CheckmarkBold    &  15k &  -  & Interactions &   Real   & - & 8 &  Normal \\
        
        RICH~\cite{RICH} &   \CheckmarkBold    &  -     &     -        &     \CheckmarkBold   & 577k & \CheckmarkBold  &  Interactions &  Real   & 142 & 22 & Normal \\
        
        AGORA~\cite{AGORA} &  \CheckmarkBold      &      - &    -         &     \CheckmarkBold   & 106.7k &  \CheckmarkBold &  Daliy &  Synthetic   & - & - & - \\
       
        BEDLAM~\cite{black2023bedlam} &  \CheckmarkBold     &     -  &   -         &     -   & 1M &  \CheckmarkBold &  Daliy &  Synthetic  & - & - & - \\

        FreeMotion~\cite{ren2024livehps} &  \CheckmarkBold     &     IMU  &   \CheckmarkBold          &    \CheckmarkBold  & 578k &  - &  Daliy &  Real  & - & - & Normal \\
        HmPEAR~\cite{lin2024hmpear} &  \CheckmarkBold     &     -  &   \CheckmarkBold   &   -  & 300k &  - &  Daliy &  Real  & 6k & 25 & Normal \\
        RELI11D~\cite{yan2024reli11d} &  \CheckmarkBold     &    IMU  &   \CheckmarkBold   &   \CheckmarkBold  & 239k &  \CheckmarkBold &  Sport &  Real  & 48 & 10 & Normal \\
        HiSC4D~\cite{dai2024hisc4d} &  \CheckmarkBold     &    IMU  &   \CheckmarkBold   &   \CheckmarkBold  & 36k &  \CheckmarkBold &  Sport &  Real  & 8 & - & Normal \\

        LiDARHuman51M~\cite{zhang2024lidarcapv2} &  \CheckmarkBold     &    IMU  &   \CheckmarkBold   &   -  & 374k &  - &  Daily &  Real   & 52 & 10 & Normal \\
        
        % TRUMANS~\cite{jiang2024scaling} &  \CheckmarkBold     &     IMU  &   \CheckmarkBold          &    \CheckmarkBold  & 1.6M &  \CheckmarkBold  &  Daliy &  Synthetic  & 7 & 20  \\
        
        % \textcolor[rgb]{0.066,0.466,0.69}{\textbf{AscendMotion(Ours)}}
        %  &   \textcolor[rgb]{0.066,0.466,0.69}\CheckmarkBold    &   \textcolor[rgb]{0.066,0.466,0.69} {IMU}   &    \textcolor[rgb]{0.066,0.466,0.69}\CheckmarkBold       &      \textcolor[rgb]{0.066,0.466,0.69}\CheckmarkBold  & \textcolor[rgb]{0.066,0.466,0.69}{\textbf{xxx}} &  \textcolor[rgb]{0.066,0.466,0.69}\CheckmarkBold \textcolor[rgb]{0.066,0.466,0.69}{\textbf{(xx)}} &  \textcolor[rgb]{0.066,0.466,0.69}{\textbf{Climbing}} & 
        %  \textcolor[rgb]{0.066,0.466,0.69}{\textbf{xxx}} & 
        %  \textcolor[rgb]{0.066,0.466,0.69}{\textbf{xxx}} & 
        %  \textcolor[rgb]{0.066,0.466,0.69}{\textbf{xxx}} \\

        CIMI4D~\cite{yan2023cimi4d} &    \CheckmarkBold   &   IMU    &    \CheckmarkBold         &    \CheckmarkBold    &   180k   & \CheckmarkBold  & Climbing & Real  & 42 & 12 & Normal \\

        SPEED21~\cite{elias2021speed21} &    \CheckmarkBold   &    -   &    -         &        &   46k   &   & Climbing & Real  & 95 & - & - \\
        
        \midrule
        \textbf{AscendMotion(Ours)}
         &   \CheckmarkBold    &   IMU   &       \CheckmarkBold       &         \CheckmarkBold  &    \textbf{412k} &     \CheckmarkBold    \textbf{(12)} &     {\textbf{Climbing}} & 
            \textbf{Real} & 
            \textbf{220} & 
            \textbf{22} & \textbf{Skilled}\\

        \bottomrule[1pt]
        \end{tabular}%
    }
    % \vspace{-2mm}
        \caption{Comparisons with related HMR datasets. The "-" symbol indicates that it is not included in the dataset.% or is not mentioned.
}
        \vspace{-2mm}
        \label{tab:data_compare}

 \end{table*}

%% file: Secs/Method_ClimbingCap.tex
\section{Method: ClimbingCap}
\label{sec:method}

%\PAR{Notations.}Our method utilizes several key notations to describe human pose and trajectory information. The input of the sequence consists of $R_i^c$ and $P_i^w$, which respectively represent the video sequence input in the camera coordinate system $c$ and the point cloud sequence input in the global coordinate system $w$, where $i$ denotes a frame. We define the output as: the local body pose $\{\theta_i \in \mathbb{R}^{21 \times 3}\}_{t=0}^T$ and shape coefficient $\beta \in \mathbb{R}^{10}$ of the SMPL model, which capture the detailed configuration of the human body; the orientation from SMPL space to camera space, including the orientation $\{\Gamma_i^c \in \mathbb{R}^3\}_{i=0}^T$; the trajectory in world space, including the translation $\{\tau_i^w \in \mathbb{R}^3\}_{i=0}^T$, aligned with the global reference frame.

Climbing motion capture is challenging, involving poses with extreme limb extension and full-body exertion in camera coordinates. Moreover, it requires precise alignment with the rock wall in world coordinate as climbers are ascending. The ClimbingCap pipeline consists of three parts: separate coordinate decoding, post-processing, and semi-supervised training. An overview of the proposed pipeline is shown in \cref{fig:ClimbingCapFramework}. The separate coordinate decoding and post-processing parts take into account the unique challenges posed by climbing motion, which involves complex off-ground dynamics and interactions with scenes. The semi-supervised training part makes uses of the large-scale unlabeled climbing motion data to learn better a HMR model.

\subsection{Separate Coordinate Decoding}
\label{sec:network_design}

The separate coordinate decoding (SCD) stage extracts features from the RGB sequence and LiDAR point clouds, and predicts the poses in camera coordinates and the positions in global coordinates. 

\begin{figure*}[!tb]
    % \vspace{-3mm}
    \centering
    \includegraphics[width=1\linewidth]{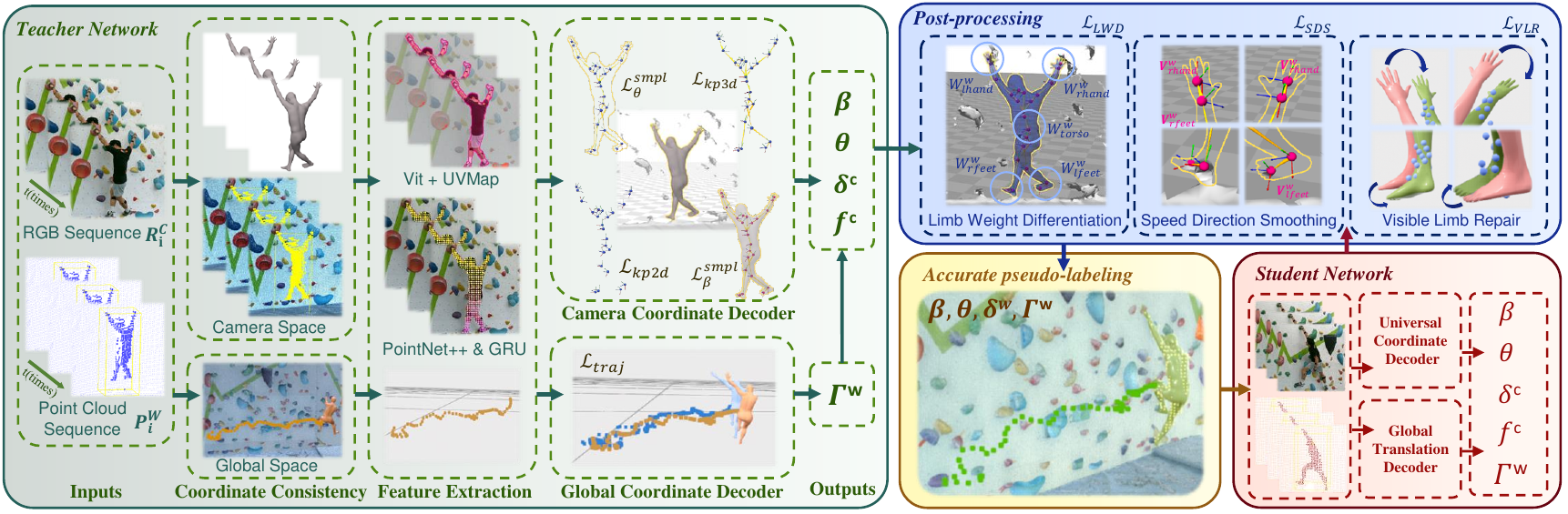}
    % \vspace{-7mm}
    \caption{\textbf{Overview of ClimbingCap.} The arrows indicate the three stages of the ClimbingCap framework: separate coordinate decoding(the \textcolor[rgb]{0.141, 0.376, 0.352}{green} box), post-processing(the \textcolor[rgb]{0.086, 0.180, 0.576}{blue} box), and semi-supervised training(the \textcolor[rgb]{0.403, 0.015, 0.066}{red} box).}  
    \vspace{-2mm}
    \label{fig:ClimbingCapFramework}
 \end{figure*}

\PAR{Input and Feature Extraction.} The overall network structure is illustrated in \cref{fig:ClimbingCapFramework}. The input includes RGB images and point cloud data. First, the point cloud data is transformed from the world coordinate system to the camera coordinate system via an extrinsic matrix, represented as
$\mathcal{P}_{\text{c}} = \Omega_{w2c} \cdot \mathcal{P}_{\text{w}}$.
Subsequently, the RGB images and transformed point cloud data are passed through feature extraction modules, \textit{RGB Extract} and \textit{PC Extract}, to obtain visual and geometric features. We build the feature extraction modules based on ViT~\cite{VIT} and PointNet++~\cite{qi2017pointnet++}. These features are then fed into the following two decoder modules, which regress the SMPL parameters and global motion parameters of the human body, respectively.

\PAR{Camera Coordinate Decoder.} This module decodes the SMPL parameters in the camera coordinate system. The RGB and point cloud features serve as inputs to the \textit{Camera Coordinate Decoder} (denoted as $\mathcal{T}_{\text{Decoder}}$), which processes the inputs with contextual information $\mathbf{f}_{\text{backbone}}$, generating an output token $\mathbf{t}_{\text{out}}$. This output token is then used to iteratively optimize the SMPL parameters, including the pose $\theta$, shape $\beta$, and camera translation $\Delta c$. The iterative decoding approach allows the model to gradually approximate the true pose and shape in the camera coordinate system. In each iteration, the decoder updates the current SMPL parameters $\theta_i$, $\beta_i$, and $\Delta c_t$ as follows:
\begin{equation}
\mathbf{t}_{\text{out}} = \mathcal{T}_{\text{Decoder}}(\mathbf{t}, \mathbf{f}_{\text{backbone}}),
\end{equation}

where $\theta_{i+1} = \Phi_{\theta} \cdot \mathbf{t}_{\text{out}} + \theta_i$, $\beta_{i+1} = \Phi_{\beta} \cdot \mathbf{t}_{\text{out}} + \beta_i$, and $\Delta c_{i+1} = \Phi_{c} \cdot \mathbf{t}_{\text{out}} + \Delta c_i$ are the update equations, with $\Phi_{\theta}$, $\Phi_{\beta}$, and $\Phi_{c}$ representing the respective weight matrices for each parameter. Here, the input token $\mathbf{t}$ can include initialized pose, shape, and camera parameters as needed.

\PAR{Global Coordinate Decoder.} To fully capture the human motion trajectory in the world coordinate, we design the \textit{Global Translation Decoder} to predict the global translation parameters $\Gamma^{\text{trans}}$ of the human body. In this module, the decoder processes the features $\mathbf{f}_{\text{backbone}}$ as contextual input, iteratively updating the global translation parameters. The update formula in each iteration is given by:
\begin{equation}
\Gamma_{i+1}^{\text{trans}} = \Psi \cdot \mathbf{t}_{\text{out}} + \Gamma_{i}^\text{trans},
\end{equation}

where $\Gamma_{i}^\text{trans}$ represents the global translation parameters at time step $i$, and $\Psi$ is the weight matrix for the update. This decoding process enables the model to capture a complete motion trajectory in the global coordinate system.

% To ensure accurate predictions of the global motion trajectory, we introduce a global trajectory loss $\mathcal{L}_{\text{traj}}$, which measures the discrepancy between the predicted translation parameters $\Gamma_{i}^\text{trans}$ and the ground truth trajectory $\Gamma^\text{transGT}$. Assuming the ground truth trajectory is $\Gamma^\text{transGT}$, the global trajectory loss is defined as:

% \begin{equation}
% \mathcal{L}_{\text{traj}} = \frac{1}{T} \sum_{t=1}^{T} \left\| \Gamma_{\text{trans}, t} - \Gamma_{\text{trans}, t}^{\text{GT}} \right\|_2^2
% \end{equation}

% where $T$ denotes the total number of time steps, and $\|\cdot\|_2$ represents the L2 norm. This loss term guides the network during training to generate smooth and accurate global motion trajectories, thereby enhancing the stability and precision of human motion capture.

\PAR{Loss.} The total loss function not only includes the 3D keypoint loss $\mathcal{L}_{\text{kp3d}}$ and 2D keypoint loss $\mathcal{L}_{\text{kp2d}}$ but also incorporates the SMPL parameter loss $\mathcal{L}^{\text{smpl}}$ and the global trajectory loss $\mathcal{L}_{\text{traj}}$. Specifically, the 3D keypoint loss $\mathcal{L}_{\text{kp3d}}$ measures the 3D error of the predicted keypoints, while the 2D keypoint loss $\mathcal{L}_{\text{kp2d}}$ measures the 2D projection error. The SMPL parameter loss $\mathcal{L}^{\text{smpl}}$ supervises the decoded pose and shape parameters, and the global trajectory loss $\mathcal{L}_{traj}$ constrains the translation parameters to within close distance of ground truth positions. Please see the supplementary for details. The final total loss is formulated as:
\vspace{-4mm}
\begin{equation}
	\begin{split}
        \mathcal{L}=
          \mathcal{L}_{kp3d} +
          \mathcal{L}_{kp2d} +
        \mathcal{L}_\theta^{smpl}
        +   \mathcal{L}_\beta^{smpl}
        +   \mathcal{L}_{traj}.
    \end{split}
\end{equation}

% \vspace{1mm}
% \begin{equation}
% 	\begin{split}
%         \mathcal{L}=
%         w_{3D} \cdot \mathcal{L}_{kp3d} +
%         w_{2D} \cdot \mathcal{L}_{kp2d} +
%         w_{\theta} \cdot 
%         \mathcal{L}_\theta^{smpl}
%         + w_{\beta} \cdot \mathcal{L}_\beta^{smpl}
%         + w_{traj} \cdot \mathcal{L}_{traj}
%     \end{split}
% \end{equation}

% where the weights $\{w_{3D}, w_{2D}, w_{\theta}, w_{\beta}, w_{\text{traj}}\}$ can be adjusted according to specific needs to achieve the best effect in subsequent pseudo-label generation and student network tranining.

\subsection{Post-processing}

Researches~\cite{shin2024wham,shen2024gvhmr,ye2023slahmr} have shown that a post-processing stage can be used to improve the output motion recovery results. Following these approaches, we employ a post-processing stage to optimize the output pose from SCD~\ref{sec:network_design} stage.  %In this stage, $\theta_t$, $\Gamma_t^w$, and $\tau_t^w$ are optimized through using the Adam~\cite{kingma2015adam} optimizer.

%\PAR{post-processing.} As mentioned in \cref{sec:coordinate_consistency}, 

One distinct advantage of ClimbingCap is that the output results for the pose decoding stage can be rigidly transformed between the camera coordinate system and the world coordinate system. Thanks to the LiDAR modality, the point cloud contains 3D information in the world coordinate system.  The poses obtained from the SCD stage are converted from the camera coordinate system to the world coordinate system through the inverse extrinsic matrix $\Omega_{w2c}^{-1}$.

The post-processing stage consists of three losses: $\mathcal{L}_{LWD}$, $\mathcal{L}_{SDS}$, and $\mathcal{L}_{VLR}$. $\mathcal{L}_{LWD}$ assigns different weights to the vertices of different parts of the climbing human SMPL, and optimizes the position of the human body in the world coordinate system according to these weights. $\mathcal{L}_{SDS}$ optimizes joint positions by smoothing the changes in the direction of the human body's velocity on the scene (i.e., rock wall). $\mathcal{L}_{VLR}$ optimizes the difficult-to-estimate limb end poses by using the position of the point cloud in space. In this stage, the global poses are optimized through using the Adam~\cite{kingma2015adam} optimizer. Please refer to the supplementary for detailed definition of these losses.%For details, see supplementary.

\subsection{Semi-supervised Training}

Compared to ground-based motions~\cite{AMASS_ICCV2019}, the size of labeled climbing motions is small. Simply using labeled climbing motion data may not be enough to train a robust model. Different from labeled climbing motion data, collecting unlabeled climbing motion data is cheaper. The AscendMotion dataset contains more unlabeled data than label data. They can be used to further improve the HMR model. 

It has been shown by researches~\cite{wang20213dioumatch,zheng2021se,xia2024hinted} from the object detection community that through using a teacher-student semi-supervised training framework, the performance of object detection models can be improved. We adopt such semi-supervised training frame for HMR. In this work, we refer the model trained after the SCD and the post-processing stage as the teacher model (green box in ~\cref{fig:ClimbingCapFramework}). The student model (red box in ~\cref{fig:ClimbingCapFramework}) clones the parameters of the teacher model. During semi-supervised training, the teacher model estimate the pose labels from unlabeled motion data, and the pose label is used as pseudo-label to further train the student model. We show in the experiment section that via utilizing the semi-supervised training framework, the performance of HMR can be further improved. Please refer to the supplementary for the detailed training process. 

%The green box represents the Teacher Network trained using fully supervised data, and the red box Student Network has the same network structure and training model. It is worth noting that the Student Network can obtain precise pseudo-labels generated by the Teacher Network and post-processing from unlabeled data, resulting in a more generalized model. For details, please refer to the supplementary.

%use a teacher-student model to train the optimized high-confidence accurate pseudo-labels.

%In addition, in order to improve the robustness of the model, we also designed a semi-supervised structure to effectively utilize the results of optimization in the world coordinate system. As shown in \cref{fig:ClimbingCapFramework}, we build our framework based on the mature semi-supervised architecture in the field of LiDAR point clouds\cite{wang20213dioumatch}\cite{zheng2021se}\cite{xia2024hinted}.The green box represents the Teacher Network trained using fully supervised data, and the red box Student Network has the same network structure and training model. It is worth noting that the Student Network can obtain precise pseudo-labels generated by the Teacher Network and post-processing from unlabeled data, resulting in a more generalized model. For details, please refer to the supplementary.

%% file: Secs/Dataset_AscendMotion.tex
 \begin{figure}[!tb]
    \centering
    \includegraphics[width=1\linewidth]{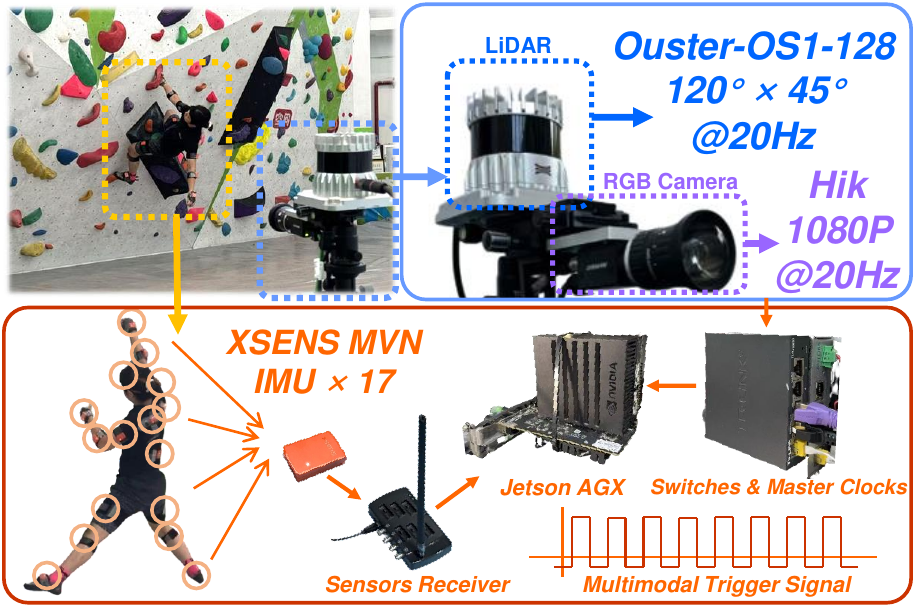}
    % \vspace{-6mm}
    \caption{Dataset collection hardware system.}
    \vspace{-2mm}
    \label{fig:Hardware}
 \end{figure}

\section{The AscendMotion Dataset}
\label{sec:dataset}

To advance research in Human Mesh Recovery (HMR) for climbing motion within a global coordinate system, we present the AscendMotion dataset. This dataset specifically captures complex multi-directional rock climbing motions on non-planar surfaces. It provides a unique opportunity for challenging HMR research. AscendMotion includes multi-modal motion data from skilled climbers performing various climbing styles, such as bouldering, speed climbing, and lead climbing, with synchronized recordings from IMU, RGB cameras, and LiDAR. The dataset comprises 344 minutes (6 hours) of labeled data and 441 minutes of unlabeled data, featuring 22 adult climbers across 12 real-world climbing scenes. All participants agreed to use their recorded data for scientific purposes. 
% Rock climbing is extremely challenging for unskilled climbers, who can easily fall off the rocks, while skilled climbers can climb rocks faster and longer, and hold onto more rock holds than amateurs.

\subsection{Hardware and Configuration}

The AscendMotion data collection system integrates multiple sensors and captures motion data in both indoor and outdoor environments. As shown in Fig. 3, the LiDAR (Ouster-OS1, 128 beams) captures 3D dynamic point clouds at 20 frames per second (FPS), while the Hik 1080P RGB camera records RGB video at 20 FPS. For each climbing environment, we use the Trimble X7 3D laser scanner to reconstruct high-resolution RGB point cloud scenes, each containing approximately 80 million points. For the labeled dataset, each climber wears an Xsens MVN inertial motion capture system with 17 IMUs, recording at 60 FPS. ~\cite{bhatnagar2020combining} and a handheld point cloud scanner to ensure accurate body shape representation. For the unlabeled dataset, climbers do not wear a MoCap outfit, and we record their climbing motions using a RGB camera and a LiDAR. A human motion is represented by $M=(T, \theta, \beta)$, where $T$ represents global translation, $\theta$ is the SMPL~\cite{SMPL2015} pose parameters, and $\beta$ is the SMPL shape parameter. We obtain each climber's shape parameters $ \beta $ using IPNet~\cite{bhatnagar2020combining}. 

 \begin{figure*}[!tb]
    % \vspace{-3mm}
    \centering
    \includegraphics[width=1\linewidth]{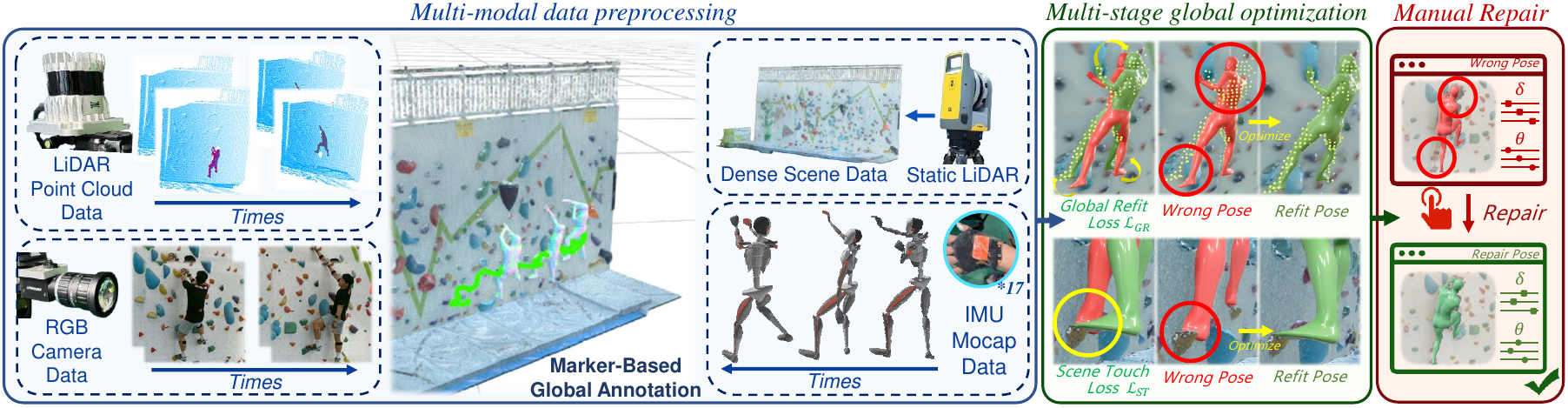}
    % \vspace{-7mm}
    \caption{\textbf{AscendMotion Annotation Pipeline.} From left to right, this pipeline represents the three stages of dataset annotation: time-space synchronous input pre-processing(the \textcolor[rgb]{0.184, 0.321, 0.560}{blue} box), multi-stage global optimization(the \textcolor[rgb]{0.043, 0.305, 0.043}{green} box), and manual repair(the \textcolor[rgb]{0.521, 0.086, 0.109}{red} box).}  
    % \vspace{-4mm}
    \label{fig:AscendMotionPipeline}
 \end{figure*}

\subsection{Annotation Pipeline}

The pose $\theta$ and the translation $T$ obtained from the IMU measurements may be inaccurate. Especially the translation $T$ may significantly drift for long-duration capture. The annotation pipeline of AscendMotion is used to find the accurate translate $T$ and pose $\theta$ as labels. 

The annotation pipeline of AscendMotion is depicted in~\cref{fig:AscendMotionPipeline}. It consists of three parts: preprocessing, multi-stage global optimization, and manual annotation. The preprocessing part is used for time synchronization and spatial calibration. The multi-stage global optimization part is developed for automatic generation of annotation labels. The manual annotation part is used to further improve the quality of generated annotations.

% To ensure the quality of the dataset, we implemented an automatic annotation pipeline that utilizes the motion characteristics inherent to climbing, such as spatial consistency and contact dynamics, to refine pose and trajectory annotations frame by frame. We further enhance annotation quality through manual verification and correction processes, as described in \cref{sec:DataPipeline}. In \cref{sec: Dataset Evaluations}, we conduct qualitative and quantitative evaluations of AscendMotion, validating its efficacy for climbing motion analysis.

\subsubsection{Multi-modal Data Preprocessing Stage}

\PAR{Time Synchronization.} The time among RGB camera and LiDAR is synchronized via Precision Time Protocol (PTP). We employ CollShark Auto 66 unit as the master clock, and it sends PTP slave clocks to the RGB camera and LiDAR. The time of IMU MoCap is post-synchronized with the LiDAR and RGB through anchor frames. 

\PAR{Calibration.} First, the LiDAR point cloud are registered with high-precision scanned-scenes. Next, the coordinate of LiDAR is treated as the world coordinate. The IMU measurements are transformed into the world coordinate through a calibration matrix. Finally, we perform frame-level calibration among RGB, LiDAR and IMU. Please refer to the appendix for details.

\subsubsection{Multi-stage Global Optimization}
\label{sec:Multi-stage Global Optimization}

AscendMotion uses the translation $T$ and pose $\theta$ provided by the IMU MoCap as the initialization of annotation labels, and performs multi-stage global optimization. To achieve accurate and natural human motion data consistent with the scene, we apply two loss functions: the Global Refit Loss $ \mathcal{L}_{\text{GR}} $ and the Scene Touch Loss $ \mathcal{L}_{\text{ST}} $. 

\PAR{Global Refit Loss $\mathcal{L}_{\text{GR}}$.} The global refit loss promotes precise global alignment between the SMPL model and the point clouds. $\mathcal{L}_{\text{GR}}$ calculates the geometric discrepancy between the SMPL model vertices $V$ and the human body point cloud $P$ using a modified Chamfer distance. It applies different loss terms ($f(|v_i - p_j|^2)$ or $g(|v_i - p_j|^2)$) to different body parts to make body regions, such as the torso and limbs, match the point clouds of the human. We calculate the Euclidean distance between a vertex $v_i \in V$ and a point $p_j \in P$ as $d_{v_i,p_j}$. A loss term $f(|v_i - p_j|^2)$ is applied to a vertex $v_i$ if $d_{v_i,p_j} \leq d_{\text{torso}}$, and $g(|p_j - v_i|^2)$ is applied to $v_i$ if $d_{v_i,p_j} \leq d_{\text{limb}}$, where $d_{\text{torso}}$ and $d_{\text{limb}}$ are the distance thresholds for torso and limbs, respectively. During climbing, humans use limbs to climb up, the distance among limbs and scenes should smaller than the distance among torso and scenes, $d_{\text{limb}} \leq d_{\text{torso}}$.

\PAR{Scene Touch Loss $\mathcal{L}_{\text{ST}}$.} This loss prevents unrealistic intersections between the SMPL model and the scene mesh by measuring the penetration depth between the SMPL vertices $v_i \in V$ and the scene mesh vertices $q_j \in Q$. Given the scene mesh normal vectors $n_j$, the penetration depth for each vertex is calculated as the dot product $\eta(v_i) = (v_i - q_j) \cdot n_j$, where $q_j$ is the closest mesh vertex to $v_i$. If the penetration depth $\eta(v_i)$ is negative, it indicates that the vertex $v_i$ has penetrated the scene mesh, and this value contributes to the loss $\mathcal{L}_{\text{ST}}$. This loss encourages the SMPL model to avoid intersecting with the scene mesh, promoting more realistic interactions between the model and the environment.

%\PAR{Scene Touch Loss $\mathcal{L}_{\text{ST}}$:} This loss prevents unrealistic intersections between the SMPL model and the scene mesh by measuring the penetration depth between the SMPL vertices $v_i \in V$ and the scene mesh vertices $q_j \in Q$. Given the scene mesh normal vectors $n_j$, the penetration depth for each vertex is calculated as the dot product $\eta(v_i) = (v_i - q_j) \cdot n_j$, where $q_j$ is the closest mesh vertex to $v_i$. If the penetration depth $\eta(v_i)$ is negative, it indicates that the vertex $v_i$ has penetrated the scene mesh, and this value contributes to the loss. The indicator function $\mathbb{I}(\eta(v_i) < 0)$ ensures that only those vertices that have penetrated the mesh contribute to the loss. This loss encourages the SMPL model to avoid intersecting with the scene mesh, promoting more realistic interactions between the model and the environment. For further details, please refer to the supplementary.

\subsubsection{Manual Annotation and Verification}

%After the multi-stage global optimization phase, the generated 

To further improve the dataset quality, we use the SMPL annotation tool~\cite{wangpedestrian} for enhanced labeling of sequences. We engage four observers to review the automatically annotated results. For any ambiguous limb movements observed in RGB views, we perform additional manual adjustments and mark these frames as key frames. Finally, we propagate poses from multiple key frames within each sequence to ensure global consistency in both smoothness and accuracy.

%% file: Secs/Experiment.tex
\begin{table}[!tb]
    % \vspace{-3mm}
    \centering
	\resizebox{\linewidth}{!}{
    \begin{tabular}{cc|ccc|ccc}
    \toprule
    \multicolumn{2}{c|}{Constraint term} & \multicolumn{3}{c|}{Horizontal Scene} & \multicolumn{3}{c}{Vertical Scene} \\
    
    $\mathcal{L}_{ST}$ & $\mathcal{L}_{GR}$   & ACCEL$\downarrow$ & MPJPE$\downarrow$ & PA-MPJPE$\downarrow$ & ACCEL$\downarrow$ & MPJPE$\downarrow$  & PA-MPJPE$\downarrow$   \\
    \midrule
    \textcolor{red}{\XSolidBrush} & \textcolor{red}{\XSolidBrush} &  1.83 & 37.57 & 32.52 & 6.79  & 51.51 & 46.24  \\
    
    \textcolor{green}{\Checkmark} & \textcolor{red}{\XSolidBrush} & 1.67 & 35.45 & 31.81 &  3.73   & 40.36 & 35.37   \\
    
    \textcolor{red}{\XSolidBrush} & \textcolor{green}{\Checkmark} &  1.59 & 29.04 & 25.95 &  1.66  & 29.24 & 24.99  \\

    \textcolor{green}{\Checkmark} & \textcolor{green}{\Checkmark} &  1.54 & 28.20 & 23.25 & 1.60  & 29.07 & 24.37  \\

    \bottomrule
    \end{tabular}%
    }
    % \vspace{-3mm}
    \caption{Quantitative Evaluation of Dataset Quality.}
    % \vspace{-3mm}
    \label{tab:eva_AscendMotion}
\end{table}

\input{tables/benchmark_in_AscendMotion_CIMI4D}

\section{EXPERIMENTS}

In this section, we demonstrate the high-quality of the AscendMotion optimization process in ~\cref{sec:exp:dataset}, show that the AscendMotion dataset is challenging to existing HMR methods, and evaluate ClimbingCap and multiple HMR methods in climbing motion datasets in Sec.~\ref{sec:exp:methods}. We demonstrate the promising performance of ClimbingCap and shows that each component of ClimbingCap is useful through ablation study in~\cref{sec: Ablation}. Please refer to the detailed experimental setup and metrics in the supplementary. 

%\PAR{Configuration and Metrics.} 
%The metrics in camera coordinates include the widely used MPJPE, Procrustes aligned MPJPE (PA-MPJPE), per vertex error (PVE), acceleration error (Accel, in units of $m/s^2$), and relative global human root translation error (T-Error in $m$).

%We follow the global and local human evaluation protocols of~\cite{shin2024wham}~\cite{shen2024gvhmr} and the global human trajectory evaluation protocol of~\cite{dai2023sloper4d}~\cite{yan2024reli11d}. We use the released code of GVHMR and SLOPER4D for testing and unify the error units. As mentioned in the test of WHAM, to calculate the world coordinate metrics, the predicted global sequence is divided into segments of 100 frames, and each segment is aligned with the ground truth segment as follows. When the alignment is performed using the entire segment, we report the world aligned average per-joint position error (WA-MPJPE). When the alignment is performed using the first two frames, we report the world MPJPE (W-MPJPE). In addition, to evaluate the error of global motion, we evaluate the root translation error (RTE in $\%$) and motion jitter (Jitter, in $m/s^3$) of the entire sequence. The metrics in camera coordinates include the widely used MPJPE, Procrustes aligned MPJPE (PA-MPJPE), per vertex error (PVE), acceleration error (Accel, in units of $m/s^2$), and relative global translation error (T-Error in $m$).

\subsection{Datasets Evaluation}\label{sec:exp:dataset}

To quantitatively evaluate the annotation quality of AscendMotion, we divide the scene into horizontal and vertical rock walls and evaluate the performance of the global optimization stage (in ~\cref{sec:Multi-stage Global Optimization} ) by comparing the generated annotations with manual annotations. 

The evaluation metrics include widely used MPJPE, Procrustes Aligned MPJPE (PA-MPJPE), per vertex error (PVE), acceleration error (Accel, in units of $m/s^2$). These metrics are used for evaluating HMR methods as well. 

To understand the impact of different constraints used in the combined optimization stage, we perform an ablation study on two different losses: $\mathcal{L}_{ST}$ and ${\mathcal{L}}_{GR}$. ~\cref{tab:eva_AscendMotion} shows the error metrics using different loss combinations for two scenes. The error metrics are small, demonstrating the effectiveness of the global optimization pipeline and the high quality of AscendMotion.

%To quantitatively evaluate the annotation quality of AscendMotion, we divide the scene into horizontal and vertical rock walls and evaluate the performance of the global ptimization stage (in ~\cref{sec:Multi-stage Global Optimization} ) by comparing the generated annotations with manual annotations. The last row of \cref{tab:eva_AscendMotion} describes the error metrics without/with the annotations generated in the optimization stage. The error metrics are small, demonstrating the effectiveness of the annotation pipeline and the high quality of AscendMotion. To understand the impact of different constraints used in the combined optimization stage, we perform an ablation study on two different losses: $\mathcal{L}_{ST}$ and ${\mathcal{L}}_{GR}$. ~\cref{tab:eva_AscendMotion} shows the error metrics using different loss combinations for the two scenes. The error metrics are the largest when no loss is used (row 1). If we remove any loss from the optimization stage (rows 2-3), the error metric increases, indicating that all losses help improve the quality of our dataset. 

\subsection{Comparison on Global Motion Recovery}\label{sec:exp:methods}

%We use the released code of GVHMR and SLOPER4D for testing and unify the error units.

To evaluate the performance of global HMR, we follow the motion evaluation protocols of~\cite{shin2024wham,shen2024gvhmr} and the global trajectory evaluation protocol of~\cite{dai2023sloper4d,yan2024reli11d}. Following WHAM~\cite{shin2024wham}, to calculate the world coordinate MPJPE metrics, the predicted global sequence is divided into segments of 100 frames, and each segment is aligned with the ground truth segment either to the entire segment (WA-MPJPE) or to the first two frames (W-MPJPE). Besides WA-MPJPE and W-MPJPE, the root translation error (RTE),  motion jitter (Jitter, in $m/s^3$), relative global translation error (T-Error in $m$) of the entire sequence are reported. 

We evaluate ClimbingCap against \emph{nine} methods with different modalities. The global RGB methods are TRACE~\cite{sun2023trace}, SLAHMR~\cite{ye2023slahmr}, WHAM~\cite{shin2024wham}, and GVHMR~\cite{shen2024gvhmr}. The LiDAR-based methods are LiDARCapV2~\cite{zhang2024lidarcapv2} and LiveHPS~\cite{ren2024livehps}. The LiDAR+RGB methods are ImmFusion~\cite{chen2023immfusion}, FusionPose~\cite{cong2022weakly}, and LEIR~\cite{yan2024reli11d}.

%The test results of some methods on these two types of rock walls are too different, we distinguish the evaluation results based on the test sets of these two scenes, as shown in\cref{tab:benchmark_in_AscendMotion}. 

\PAR{Results in AscendMotion.} The scenes in AscendMotion are categories into horizontal and vertical scenes, based on the major direction of human motions. Vertical motions are more challenging to human and to HMR methods than horizontal motions. The experimental results are depicted in \cref{tab:benchmark_in_AscendMotion}. The top rows depicted the results for multiple state-of-the-art RGB-based HRM methods, and the bottom rows depicted the results for state-of-the-art LiDAR and LiDAR+RGB based methods. In each cell of the table, the results for the horizontal and the vertical scenes are separated by ``/". For horizontal scenes, ClimbingCap performs the best in most of the metrics, especially in terms of MPJPE, WA-MPJPE and W-MPJPE. WHAM and GVHMR performs well for the camera coordinate metrics, but not in the world coordinate metrics. For the vertical scenes, ClimbingCap performs the best and significantly better than the second-best methods. GVHMR, a representative global HMR method, performs poorly in vertical scenes. It estimates the direction of movement by predicting the horizontal velocity. However, the major movement on the vertical scene of AscendMotion is an upward climbing movement. The LiDAR-based, and other LiDAR+RGB-based methods perform inferior to ClimbingCap. These methods do not fully consider global trajectories. The results demonstrate the importance of considering the relationship among camera coordinates and global coordinates.

\PAR{Results in CIMI4D.} To test the generalization ability of all the HMR methods, we evaluate their performance in the CIMI4D climbing dataset without retraining or fine-tuning. The CIMI4D dataset consists mostly horizontal scenes, and the motions are less challenging than AscendMotion. As it is reported in \cref{tab:benchmark_in_CIMI4D}, ClimbingCap \emph{performs the best in the CIMI4D dataset for most the metrics}, except the WA-MPJPE metric.

%In the generalization verification experiment of the method, we use the CIMI4D rock climbing dataset, which mainly consists of simple rock climbing actions. As can be seen from\cref{tab:benchmark_in_CIMI4D}, our method still has a strong adaptability to the rock climbing actions in the wild. In particular, for multimodal input, ClimbingCap can obtain the best test results in the test. As shown in Figure 4 (b), in the world coordinate system, we have accurate motion recovery capabilities for long sequences of lateral movements of rock climbing walls, and have the ability to correctly estimate human poses. Other methods, such as GVHMR with single RGB input, also achieved excellent results, because there is no rapid upward climbing action in the CIMI4D test set. The same results are also reflected in other methods.

 \begin{figure*}[!tb]
    % \vspace{-3mm}
    \centering
    \includegraphics[width=1\linewidth]{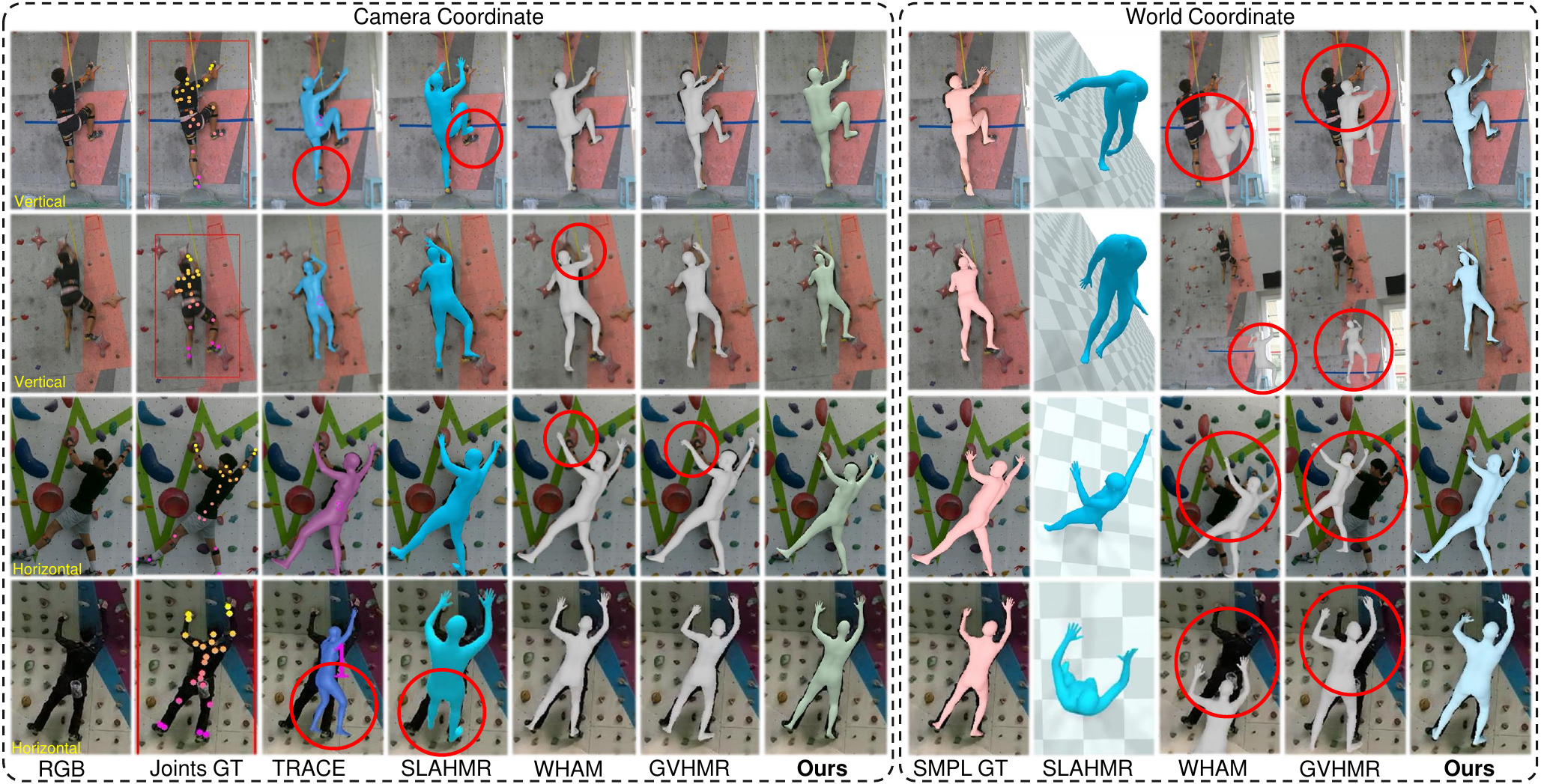}
    % \vspace{-7mm}
    \caption{\textbf{Qualitative Evaluation in the AscendMotion and CIMI4D dataset.} The left and right areas show the results of Camera Coordinate and World Coordinate respectively. The red circles indicate obvious errors. The last row shows the results for CIMI4D dataset. 
    % Comparison methods perform well in the camera coordinate system of the horizontal rock wall scene. However, due to the defect that the WHAM and GVHMR methods cannot recover the motions on the rock wall, their performance is poor. 
    Our method ClimbingCap performs best qualitatively by comparison.}  
    % \vspace{-4mm}
    \label{fig:Benchmark in dataset}
 \end{figure*}

\begin{table}[!tb]
    % \vspace{-3mm}
    \centering
	\resizebox{\linewidth}{!}{
    \begin{tabular}{lcccccc}
    \toprule
    Variant & MPJPE & PA-MPJPE & PCK0.3 & WA-MPJPE & W-MPJPE & RTE   \\
    \midrule
    (1) RGB Input  & 105.67 & 63.05 & 0.78 & 117.17 & 174.53 & 7.64   \\
    (2) w/o $\mathcal{L}_{LWD}$ & 80.46 & 52.15 & 0.89 & 70.04 & 91.23 & 2.02  \\
    (3) w/o $\mathcal{L}_{SDS}$ & 99.13 & 60.66 & 0.81  & 109.35 & 164.11 & 7.03   \\
    (4) w/o $\mathcal{L}_{VLR}$ & 91.10 & 61.85 & 0.83 & 88.59 & 120.11 & 3.34 \\
    (5) w/o $SS$ & 77.43 & 52.57 & 0.90 & 65.30 & 82.09 & 1.83 \\

    \midrule
    RGB+LiDAR Input                 & \multirow{2}{*}{\textbf{75.45}}  & \multirow{2}{*}{\textbf{50.51}} & \multirow{2}{*}{\textbf{0.91}} & \multirow{2}{*}{\textbf{62.95}} & \multirow{2}{*}{\textbf{78.99}} & \multirow{2}{*}{\textbf{1.57}}  \\
    Full Model                 &  &  &  &  & \  \\

    \bottomrule
    \end{tabular}
    }
    % \vspace{-3mm}
    \caption{Ablation Experiment for ClimbingCap in AscendMotion.}
    \vspace{-4mm}
    \label{tab:Ablation_in_ClimbingCap}
\end{table}

\subsection{Ablation Experiment}
\label{sec: Ablation}

To understand the impact of each global loss and the semi-supervised module on ClimbingCap's performance, we evaluate five variants of ClimbingCap on the AscendMotion dataset following the same training and evaluation protocol as the previous section. (1) RGB Input: Using only RGB imagery without LiDAR, this variant performs worst overall due to missing global position data required for post-processing. (2) w/o $\mathcal{L}_{LWD}$: Excluding the limb weight differentiation loss, which guides optimization using global point clouds, has a limited impact, indicating a beneficial but non-essential role. (3) w/o $\mathcal{L}_{SDS}$: Removing the velocity direction smoothing loss, which ensures motion consistency, significantly reduces global metrics, highlighting its importance. (4) w/o $\mathcal{L}_{VLR}$: The velocity direction smoothing loss constrains limb movements and corrects unexpected mis-predicted movements in a motion sequence; without this loss, ClimbingCap’s global metrics drop significantly, highlighting its importance for consistency and accuracy. (5) w/o $SS$: Removing the semi-supervised module results in a slight performance decrease, indicating its contribution to improving network training, though with a modest effect. The full ClimbingCap model, with RGB and LiDAR inputs, all loss terms, and the semi-supervised framework, achieves the best results in \cref{tab:Ablation_in_ClimbingCap}, validating the design’s effectiveness for climbing motion analysis on AscendMotion.

%% file: tables/benchmark_in_AscendMotion_CIMI4D.tex
\begin{table*}[!t]
    \centering
    \resizebox{\linewidth}{!}{

    \begin{tabular}{c|c|ccccc|ccccc}
    
    \toprule
    \multirow{2}{*}{Modality} & \multirow{2}{*}{Method} & \multicolumn{5}{c|}{Camera Coordinate} & \multicolumn{5}{c}{World Coordinate} \\
    \cmidrule(lr){3-7} \cmidrule(lr){8-12}
    & & ACCEL$\downarrow$ & MPJPE$\downarrow$ & PA-MPJPE$\downarrow$ & PVE$\downarrow$ & PCK0.3$\uparrow$ & WA-MPJPE$\downarrow$ & W-MPJPE$\downarrow$ & RTE$\downarrow$ & Jitter$\downarrow$ & T-Error$\downarrow$ \\
    \midrule

    \multirow{4}{1cm}{\centering RGB} 

    & \centering TRACE~\cite{sun2023trace} & 18.68/76.79 & 875.56/577.60 & 69.21/85.81 & 951.52/619.93 & 0.06/0.09 & 144.33/\textbf{385.71} & 254.38/703.35 & 14.73/26.17 & 115.96/521.40 & 2.56/6.62\\

    & \centering SLAHMR~\cite{ye2023slahmr} & 5.46/96.98 & 232.46/467.88 & 84.13/285.15 & 283.24/552.63 & 0.36/0.12 & 277.47/447.68 & 804.85/\textbf{613.60} & \textbf{3.64}/39.39 & 4.91/201.33 & 2.81/6.54\\
    
    & \centering WHAM~\cite{shin2024wham} & 4.59/35.01 & 110.92/143.17 & 76.09/\textbf{73.36} & 124.2/164.91 & 0.76/0.62 & 229.42/1125.77 & 647.70/1499.85 & 5.16/9.04 & \textbf{3.58}/40.69 & 1.77/\textbf{2.49}\\
    
    & \centering GVHMR~\cite{shen2024gvhmr} & \textbf{4.50}/\textbf{26.22} & \textbf{107.09}/\textbf{124.60} & \textbf{60.06}/80.30 & \textbf{118.89}/\textbf{151.10} & \textbf{0.77}/\textbf{0.71} & \textbf{105.15}/1002.11 & \textbf{202.45}/1442.50 & 4.09/\textbf{7.91} & 6.85/\textbf{32.71} & \textbf{1.48}/2.54\\
    
    % & \centering \textbf{Ours} & 7.55/32.13 & \textbf{105.67}/164.37 & 63.05/\textbf{74.56} & 119.96/185.33 & \textbf{0.78}/\textbf{0.57} & 117.17/\textbf{229.02} & \textbf{174.53}/432.88 & 7.64/\textbf{4.17} & 7.21/\textbf{39.91} & 1.89/2.55\\

    \midrule 

    \multirow{2}{1cm}{\centering LiDAR} 
    
    & \centering LiDARCapV2~\cite{zhang2024lidarcapv2} & 87.99/119.62 & 244.6/234.52 & 192.17/156.39 & 326.45/283.27 & 0.53/0.50 & 282.12/1396.42 & 442.12/1518.29 & 16.42/10.85 & 176.95/165.55 & 1.65/2.89\\
    
    & \centering LiveHPS~\cite{ren2024livehps} & 157.87/195.23 & 156.5/147.31 & 142.19/121.76 & 191.87/189.30 & 0.64/0.70 & 235.4/1369.89 & 392.34/1506.50 & 13.94/10.45 & 279.96/358.54 & 2.1/6.73\\
    
    \noalign{\vskip 2pt}
    \multirow{4}{1cm}{\centering LiDAR\\+RGB}

    & \centering ImmFusion~\cite{chen2023immfusion} & 108.76/74.20 & 473.18/464.83 & 254.07/179.51 & 533.5/529.71 & 0.17/0.14 & 324.4/1446.01 & 487.92/1532.88 & 16.52/10.86 & 27.03/\textbf{14.49} & 1.94/3.99\\
    
    & \centering FusionPose~\cite{cong2022weakly} & 112.08/86.44 & 256.81/315.93 & 198.55/193.83 & 306.22/359.68 & 0.36/0.42 & 275.02/1445.32 & 444.47/1532.28 & 16.29/10.85 & 92.97/80.79 & 2.02/6.48\\
    
    & \centering LEIR~\cite{yan2024reli11d} & 110.18/94.57 & 297.95/299.62 & 187.26/150.56 & 340.61/351.52 & 0.41/ 0.37 & 266.82/1313.09 & 282.31/1435.92 & 9.78/9.97 & 73.38/85.03 & 1.1/\textbf{1.20}\\
    
    & \centering \textbf{Ours} & \textbf{5.17}/\textbf{17.25} & \textbf{75.45}/\textbf{88.92} & \textbf{61.73}/\textbf{74.50} & \textbf{94.89}/\textbf{106.42} & \textbf{0.91}/\textbf{0.78} & \textbf{62.95}/\textbf{85.26} & \textbf{78.99}/\textbf{106.95} & \textbf{1.57}/\textbf{3.12} & \textbf{8.3}/27.75 & \textbf{1.07}/1.29 \\

     \bottomrule
     \end{tabular}%
     }

     \vspace{-1mm}
     \caption{\textbf{HMR Comparison in the AscendMotion dataset (Horizontal Scene/Vertical Scene).} The results for the horizontal and the vertical scene are separated by ``/'' in each cell. ClimbingCap performs significantly than others.
     }
     % \vspace{-2 mm}
	\label{tab:benchmark_in_AscendMotion}
\end{table*}

\begin{table*}[!t]
    \centering
    \resizebox{\linewidth}{!}{
    \begin{tabular}{c|c|ccccc|ccccc} 
    
    \toprule
    \multirow{2}{*}{Modality} & \multirow{2}{*}{Method} & \multicolumn{5}{c|}{Camera Coordinate} & \multicolumn{5}{c}{World Coordinate} \\
    \cmidrule(lr){3-7} \cmidrule(lr){8-12}
    & & ACCEL$\downarrow$ & MPJPE$\downarrow$ & PA-MPJPE$\downarrow$ & PVE$\downarrow$ & PCK0.3$\uparrow$ & WA-MPJPE$\downarrow$ & W-MPJPE$\downarrow$ & RTE$\downarrow$ & Jitter$\downarrow$ & T-Error$\downarrow$ \\
    \midrule

    \multirow{4}{1cm}{\centering RGB} 

    & TRACE~\cite{sun2023trace} & 57.65 & 488.63 & 98.83 & 685.29 & 0.17 & 365.11 & 608.38 & 28.98 & 27.30 & 4.58 \\
    & SLAHMR~\cite{ye2023slahmr} & \textbf{6.10} & 228.82 & \textbf{82.39} & 201.42 & 0.61 & 391.24 & 445.38 & \textbf{13.88} & 12.63 & 2.47 \\
    & WHAM~\cite{shin2024wham} & 9.11 & 184.43 & 94.40 & 191.27 & \textbf{0.76} & \textbf{321.58} & \textbf{430.35} & 32.95 & \textbf{9.26} & 3.78 \\
    & GVHMR~\cite{shen2024gvhmr} & 6.98 & \textbf{152.56} & 91.13 & \textbf{140.19} & 0.60 & 349.17 & 476.90 & 21.54 & 11.63 & \textbf{2.26} \\
    % & Ours & 7.71 & 127.61 & 86.59 & 124.32 & 0.74 & 308.83 & 418.01 & 19.66 & 16.34 & 2.74 \\

    \midrule 

    \multirow{2}{*}{LiDAR} 
    & LiDARCapV2~\cite{zhang2024lidarcapv2} & 70.72 & 389.97 & 267.76 & 364.26 & 0.50 & 468.00 & 520.06 & 13.56 & 54.25 & 3.12 \\
    & LiveHPS~\cite{ren2024livehps} & 72.57 & 190.86 & 148.00 & 225.65 & 0.59 & 412.22 & 524.47 & 12.11 & 50.77 & 3.52 \\

    \noalign{\vskip 2pt}
    \multirow{4}{*}{LiDAR+RGB} 
    & ImmFusion~\cite{chen2023immfusion} & 68.82 & 322.21 & 232.44 & 435.72 & 0.45 & 465.46 & 586.36 & 13.59 & 14.55 & 3.96 \\
    & FusionPose~\cite{cong2022weakly} & 58.54 & 242.20 & 189.92 & 330.09 & 0.53 & 373.46 & 437.88 & 13.54 & 48.30 & 2.01 \\
    & LEIR~\cite{yan2024reli11d} & 16.46 & 206.69 & 117.52 & 269.12 & 0.58 & \textbf{228.1} & 370.74 & 10.6 & 38.42 & 1.84 \\
    & Ours & \textbf{5.18} & \textbf{84.03} & \textbf{60.69} & \textbf{99.89} & \textbf{0.86} & 228.28 & \textbf{261.06} & \textbf{8.17} & \textbf{8.72} & \textbf{1.26} \\

     \bottomrule
     \end{tabular}%
     }

     \vspace{-1mm}
     \caption{\textbf{HMR Comparison in the CIMI4D dataset.} ClimbingCap demonstrates remarkable generalization and robustness on other climbing datasets. 
     }
     % \vspace{-3mm}
	\label{tab:benchmark_in_CIMI4D}
\end{table*}

%% file: Secs/Conclusion.tex
\section{Conclusion}
\label{Sec: Conclusion}

In this work, we propose AscendMotion, a large-scale and multi-modal climbing motion dataset, and ClimbingCap, a human motion recovery framework for climbing motions. AscendMotion provides a rich, high-quality dataset that surpasses previous climbing datasets in both scale and complexity. ClimbingCap effectively recovers 3D climbing motions in the global coordinate system, ensuring accurate pose estimation and global localization. 
Experimental results demonstrate the quality of AscendMotion dataset, and show that ClimbingCap outperforms existing methods in terms of accuracy and robustness.

\PAR{Acknowledgment.} This work was partially supported by the Fundamental Research Funds for the Central Universities (No. 20720230033);  by PDL (2022-PDL-12); by Xiaomi Young Talents Program.